# CEC: Crowdsourcing-based Evolutionary Computation for Distributed Optimization

Feng-Feng Wei, Wei-Neng Chen, *Senior Member*, *IEEE*, Xiao-Qi Guo, Bowen Zhao, *Member*, *IEEE*, Sang-Woon Jeon, *Member*, *IEEE*, and Jun Zhang, *Fellow*, *IEEE*

**Abstract**—Crowdsourcing is an emerging computing paradigm that takes advantage of the intelligence of a crowd to solve complex problems effectively. Besides collecting and processing data, it is also a great demand for the crowd to conduct optimization. Inspired by this, this paper intends to introduce crowdsourcing into evolutionary computation (EC) to propose a crowdsourcing-based evolutionary computation (CEC) paradigm for distributed optimization. EC is helpful for optimization tasks of crowdsourcing and in turn, crowdsourcing can break the spatial limitation of EC for large-scale distributed optimization. Therefore, this paper firstly introduces the paradigm of crowdsourcing-based distributed optimization. Then, CEC is elaborated. CEC performs optimization based on a server and a group of workers, in which the server dispatches a large task to workers. Workers search for promising solutions through EC optimizers and cooperate with connected neighbors. To eliminate uncertainties brought by the heterogeneity of worker behaviors and devices, the server adopts the competitive ranking and uncertainty detection strategy to guide the cooperation of workers. To illustrate the satisfactory performance of CEC, a crowdsourcing-based swarm optimizer is implemented as an example for extensive experiments. Comparison results on benchmark functions and a distributed clustering optimization problem demonstrate the potential applications of CEC.

---

## 1 INTRODUCTION

CROWDSOURCING is a technology arising with the development of social media and the Internet. It aims to obtain the work, information, or opinions from a crowd [1]. Specifically, the general paradigm of crowdsourcing includes the requester, the crowdsourcing platform and the crowd as shown in Fig. 1. There are generally four steps in crowdsourcing. 1) The requester posts a task to the crowdsourcing platform. 2) The crowdsourcing platform outsources the task to a crowd, named workers. 3) Each worker conducts the task and feedbacks information to the crowdsourcing platform via the Internet or social media. 4) The crowdsourcing platform processes the collected information and sends results to the requester. Crowdsourcing is to take the advantage of the intelligence of a crowd to address complex problems effectively, which can get solutions with high diversity and complete a large task in a shorter period of time. In the past few years, it has been developed in many variants, such as spatial crowdsourcing [2], [3], [4], team crowdsourcing [5]. and successfully applied in many collection tasks [6], [7], [8], becoming a popular technique in both business [9] and academia [10] since it emerged.

However, traditional spatial or general-purpose crowdsourcing is mainly performed for globally cross-space and large-scale information sensing. With the development of Internet of things (IoTs) [11] and explosion of data [12], edge intelligence [13] and crowd computing [14] stand out as disruptive technologies. Besides collecting information, it is much attractive and needful for the crowd to process data and make decisions. For example, in power systems, the requester posts a task to the crowdsourcing platform, which is to optimize the coordination of distributed energy resources. The crowdsourcing platform outsources the task to a crowd, which need to optimize the coordination of energy resources based on their collected power generation and consumption information [15]. In federated learning, the requester posts a task of optimizing a global model. After receiving the task from the platform, the crowd need to collect their spatial distributed data and then train their local models with the data. The platform decides to optimize a global model by integrating local models [16]. In wireless sensor networks (WSNs), distributed clustering optimization is a common demand which aims to get a globally optimized clustering results based on all collected data. When receiving the task, each worker in the crowd not only need to sense their local data but also need to optimize the local clustering integrating with some global information [17]. Considering the characteristics of cross-space and large-scale sensing, we name this kind of crowdsourcing-based distributed optimization. As the name suggesting, crowd in crowdsourcing-based distributed optimization not only need to collect information, but also need to utilize their neighbor information, even the

---

- *Feng-Feng Wei, Wei-Neng Chen and Xiao-Qi Guo are with the School of Computer Science and Engineering, South China University of Technology, Guangzhou 510006, China, with the State Key Laboratory of Subtropical Building Science, with the Guangdong-Hong Kong Joint Innovative Platform of Big Data and Computational Intelligence, South China University of Technology, Guangzhou, 510006, China, and also with the Pazhou Lab, Guangzhou, 510006, China. E-mail: fengfeng_scut@163.com, cwnraul634@aliyun.com, kallyqi@outlook.com.*
- *Bowen Zhao is with Guangzhou Institute of Technology, Xidian University, Guangzhou, 510006, China. E-mail: bwinzhao@gmail.com.*
- *Sang-Woon Jeon and Jun Zhang are with the Department of Electronics Engineering and Communication Engineering, Hanyang University, Ansan 15588, South Korea. E-mail: sangwoonjeon@hanyang.ac.kr, junzhang@ieee.org.*

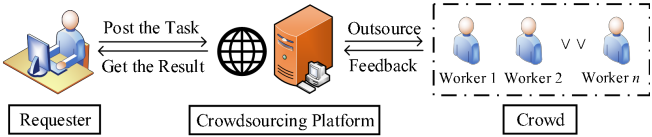

Fig. 1. The illustration of crowdsourcing.

platform or requester information to conduct optimization. With the bursting of mobile crowd sensing and computing (MCSC) [14], more applications in both industry and academia essentially belong to crowdsourcing-based distributed optimization.

Evolutionary computation (EC), including evolutionary algorithms (EAs) [18], [19] and swarm intelligence (SI) [20], [21], is a flourishing research realm for complex optimization. On the one hand, EC algorithms are a kind of derivative free optimization method, which is powerful and robust for complex optimization problems with characteristics like nonconvex, large-scale, black-box, expensive, etc. [22], [23], [24]. On the other hand, EC has intrinsic parallelism, which gives it great potential for distributed optimization by taking advantage of the power of crowd. Some researchers have studied to use EC for distributed optimization. For example, Xu *et al*. [25], [26] proposed federated data-driven EC algorithm to train the global model with distributed local models trained by distributed data. Liu *et al*. [27] designed a distributed surrogate-assisted EC for the optimization of high-dimensional feature selection. Based on distributed data and evaluation, Wei *et al*. [28] developed a distributed EC with on-demand evaluation for expensive optimization with many constraints. Guo *et al*. [29] built an edge-cloud EC framework to handle distributed data. Though many efforts have been made on EC for distributed optimization, there remain some limitations.

The first limitation is that EC suffers from *the cruse of dimensionality* when optimizing large-scale problems. *The curse of dimensionality* is a well-known challenge in EC, in which EC efficiency and optimization performance are decreasing when solving large-scale problem [30]. Though many efforts have been made to alleviate this problem through multiple processors or supercomputing techniques [31], it still a major problem to be solved when applying EC in very large-scale optimization.

The second limitation is that most existing EC for distributed optimization cannot perform well with limited data. In the literature, most studies about EC focus on centralized optimization, where global information is usually shared by all processors [31]. But in distributed optimization, information is owned by local workers and they need to communicate with each other through a specific cooperation mechanism. Although a few researchers have made attempts to apply EC on data distributed optimization problems [25], [26], [28], [29], they only considered the problem with several workers. How to achieve effective performance for large-scale problems within limited data and a large number of workers remains to be studied.

The third limitation is that EC is difficult to evolve without accurate fitness evaluation. EC conducts iterative evolution based on the principle of *survival of the fittest*, in which the fitness evaluation is specifically known [18], [19], [20], [21]. However, in distributed optimization, the heterogeneity of devices and worker behaviors may lead to uncertainties and reduce the quality of data [32]. We refer to this kind of uncertainties as the environmental-aware uncertainty. The environmental-aware uncertainty has great influence on fitness evaluations, which may mislead the evolution direction and deteriorate the algorithm performance. When designing EC in distributed environment, it is worth to study how to alleviate the negative effect led by the environmental-aware uncertainty.

Fortunately, it is promising to introduce crowdsourcing paradigm into EC to overcome the above limitations. EC and crowdsourcing are similar in the aspect that they both use lots of individuals to cooperatively complete a global task. Differently, EC is to do optimization whereas crowdsourcing is mainly used to collect and process data. There are some advantages to combine crowdsourcing with EC. Firstly, dispatching the optimization task to the crowd to inherit the paradigm of crowdsourcing is beneficial for EC to efficiently and effectively solve the problem. The original large task is cooperatively processed by lots of workers in the crowd, which is helpful to alleviate *the curse of dimensionality*. Secondly, data gained and processed by the crowd for the global interested area can provide data support for EC. It is helpful to break the limitation of data missing when apply EC for large-scale and complex distributed optimization. Thirdly, the large crowd can pass the uncertainty detection to alleviate the influence of environmental uncertainties. The crowd in crowdsourcing is generally large. If there exist workers having high-level environmental-aware uncertainties, they can be detected and removed from the evolution crowd, without influencing the whole evolution. As a result, the data and evaluation are more accuracy, and the algorithm performance can be improved as a whole. Taking the above consideration, we intend to combine EC with crowdsourcing for crowdsourcing-based distributed optimization.

Therefore, this paper introduces crowdsourcing-based distributed optimization and proposes the crowdsourcing-based EC (CEC) for it. The contributions of this paper are as follows.

Firstly, the paradigm of crowdsourcing-based distributed optimization is introduced. Similar with crowdsourcing, there are three roles, the requester, the cloud server, and workers. The requester posts the optimization task to the cloud server. The cloud server dispatches it to a group of workers. Each worker represents one individual, while all workers form a whole evolution population. Evaluations of different workers have different levels of environmental-aware uncertainties. Workers can communicate with connected neighbors and the cloud server.

Secondly, the crowdsourcing-based EC, CEC, is proposed for the above distributed optimization. It inherits the paradigm of crowdsourcing and takes the advantage of EC. Workers search for promising solutions through EC optimizer and cooperate with connected neighbors. Besides, they can communicate with the cloud server to send local information and get global information for evolution. In a word, the problem is optimized through guidance of the cloud server and cooperation of workers.

Thirdly, an uncertainty detection strategy is proposed to detect uncertainties. Specifically, connected workers can exchange fitness values of candidate solutions with environmental-aware uncertainties. They compare the received fitness with their own fitness values and send comparison results to the cloud server. The cloud server ranks all workers through the competitive ranking strategy. It detects workers with larger uncertainties according to the change of ranks. Besides, the ranking results are also used to guide the evolution of workers.

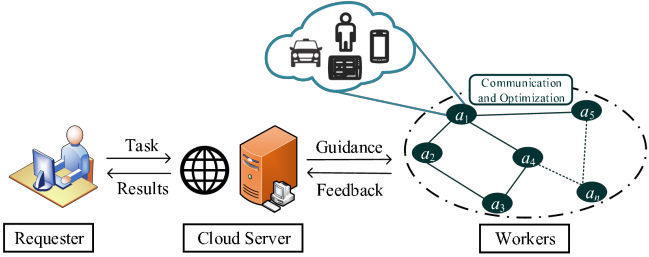

Fig. 2. The illustration of crowdsourcing-based distributed optimization, in which $a_i$ represents the $i^{th}$ worker agent.

An example, crowdsourcing-based swarm optimizer, CLLSO is implemented to illustrate the promising application of CEC. In CLLSO, the adopted EC algorithm is level-based learning swarm optimizer (LLSO), which is powerful for complex and large-scale optimization [33]. Extensive experiments on benchmark functions and a distributed clustering optimization problem demonstrate that CEC has satisfactory performance for crowdsourcing-based distributed optimization.

The rest of this paper is organized as follows. Section 2 introduces the paradigm of crowdsourcing-based distributed optimization. Section 3 elaborates the proposed CEC in detail. Section 4 conducts extensive experiments on benchmark test suites and distributed clustering optimization to investigate the performance and promising application of CEC. Section 5 concludes the whole paper.

## 2 CROWDSOURCING-BASED DISTRIBUTED OPTIMIZATION

Crowdsourcing-based distributed optimization arises with the optimization demand in crowdsourcing, in which the optimization involves lots of spatially distributed data and need to be optimized in combination with the crowd. Therefore, crowd not only need to collect information but also need to conduct optimization and decision making in multiple loops to search for the optima. Without loss of generality, the mathematical formulation of a minimization optimization problem can be shown as follows:

$$\min F = f(X, \Omega) \quad (1)$$

where $F$ is the objective to be optimized, $f$ is the pure objective evaluation, $X$ is the decision variable vector and $\Omega$ represents the set of environmental data.

As the name suggesting, crowdsourcing-based distributed optimization inherits the paradigm of crowdsourcing. Specifically, the paradigm of the crowdsourcing-based distributed optimization is shown in Fig. 2, including a requester $R$, a cloud server $S$, which is known as crowdsourcing platform in crowdsourcing and a group of workers $W$, which is also known as the crowd. Some widely used notations and the mapping between the process and mathematical formulation are shown in Table 1. The paradigm is described in detail as follows.

TABLE 1
NOTATIONS AND MAPPING BETWEEN THE PROCESS AND FORMULATION

| Notation | Description | Responsibility |
|---|---|---|
| $R$ | requester | 1) Post $P$ |
| $S$ | cloud server | 1) Outsource $P$ to $W$<br>2) Guide Search of $W$<br>3) Feedback $G$ to $R$ |
| $W$ | worker set | 1) Evolve $W$ to search<br>2) Communicate with $S$ |
| $A$ | worker agents | / |
| $E$ | edge matrix | |
| $P$ | task | |
| $X$ | decision variable | |
| $\Omega$ | environmental data | |
| $D$ | search domain | |
| $F$ | objective | |
| $f(X_{best})$ | global best fitness | |
| $X_{best}$ | global best variable | |
| $\mathbf{M}$ | comparison result matrix | |

### 2.1 The Requester

The requester $R$ is the task publisher. At the start of the crowdsourcing-based optimization, $R$ posts an optimization task, denoted by $P = \langle X, D, F \rangle$, to the cloud server to complete, where

- $X = \{x_1, x_2, ..., x_m\}$ is a decision variable vector and $m$ is the number of dimensionalities.
- $D = \{D_1, D_2, ..., D_m\} = \{[lb_1, ub_1], ..., [lb_m, ub_m]\}$ is the set of domains for each dimension. $lb_i$ and $ub_i$ ($i = 1, 2, ..., m$) represent the lower bound and upper bound of the $i^{th}$ dimension.
- $F = f(X, \Omega)$ is the objective to be optimized, which may have mathematical formulation or not. It can have complex characteristics such as nonconvex, large-scale, constrained, multi-objective, multi-modal, expensive, and so on.

The demand of the requester is asking the cloud server to feedback the found-best objective value $f(X_{best})$ and the corresponding decision variable vector $X_{best}$.

### 2.2 The Cloud Server

The cloud server $S$ is the institution linking with the requester and workers. It mainly has two functions.

1) Task Outsource. After receiving $P$ from the requester, the cloud server $S$ outsources it to the workers since the completion of the task involves lots of spatially distributed information or it is too large and complex for the cloud server to complete.

2) Search Guidance. As an institution to coordinate workers, $S$ has higher security and is believed by all workers. It participates in the optimization processes of all workers. Workers can send their optimized information to $S$. After receiving information from workers, $S$ processes it and feedback utilized global information to workers to further guide their search directions.

At the end of the optimization, the cloud server handles all optimized data and feedback $f(X_{best})$ and $X_{best}$ to the $R$.

## 2.3 Workers

Workers $W = \langle A, E \rangle$ are main task performers and have communication topology, where

- $A = a_1, a_2, ..., a_n$ is a group of worker agents and $n$ is the total number of workers. Each worker has one individual.
- $E = \{e_{ij}\}$ is the edge matrix of the crowdsourcing network, in which $e_{ij} = 1$ means worker $a_i$ is connected with worker $a_j$ and $e_{ij} = 0$ means worker $a_i$ and worker $a_j$ are not neighbors.

Workers can be vehicles, people with mobile phones, or laptops, which have small computational ability and limited storage space. Therefore, each worker has one individual in the whole search space. Apart from data collection and environmental sensing, they also have the little ability of evaluation and optimization. Out of the uncertainty of worker behaviors and heterogeneity of devices, workers have different levels of environmental-aware uncertainties. Each evaluation of the $i^{th}$ worker is affected by an uncertainty $r_i$, in which $F(X_i) = f(X_i) + r_i$. After receiving the task from the cloud server, workers cooperatively optimize it mainly through two operations, communication and optimization.

1) Communication. Due to the limited local information, workers need to communicate with their connected neighbors and the cloud server for information exchange and update. It should be noticed that workers are sensitive to raw data. As for communication with connected neighbors, they only send their environmental-aware uncertainty fitness value $F(X_i)$. As for communication with $S$, they only send the comparison results with neighbors. For each worker, its decision variable vector is kept secret from other workers unless it is required to send by a higher security institution, such as $S$.

2) Optimization

To complete the optimization task, workers need to conduct specific optimization methods, which depend on characteristics of $P$. During the optimization process, workers communicate with neighbors for information interaction. Besides local information, workers also communicate with $S$ to get partially global information for guiding the search direction.

In particular, workers are mobile and therefore their connected neighbors are randomly time-varying.

## 2.4 Characteristics

Crowdsourcing-based distributed optimization is greatly different from general-purpose distributed optimization. Taking (1) for example, as for general-purpose distributed optimization, $\Omega$ are already known. It only needs to conduct the optimization $\min f$ based on $\Omega$. The evaluation is clearly given and most studies on general-purpose distributed optimization mainly focus on the optimization mechanism. There is no data collection and they generally do not consider characteristics such as environmental-aware uncertainties, time-vary topology, data sensitivity.

Differently, in crowdsourcing-based distributed optimization, $\Omega$ are unknown. It needs to conduct crowdsensing to collect data and evolve the population for evolution at the same time. In crowdsourcing-based distributed optimization, the evaluation of each worker depends on data and environment, which may be affected by environmental-aware uncertainties, changing topology and collected data. Therefore, when designing algorithms for crowdsourcing-based distributed optimization, characteristics as environmental-aware uncertainties, weak centralization, time-vary topology and data sensitivity should be seriously considered.

From the above description, characteristics of crowdsourcing-based distributed optimization can be concluded as follows.

i) **Uncertainty fitness**. The evaluation of each worker has an environmental-aware uncertainty. This is one form of effects brought by crowdsensing. It should be considered when designing algorithm since crowdsourcing-based distributed optimization simultaneously conduct data sensing and optimization.

ii) **Weak centralization**. Workers can communicate with the cloud server to send partially local information and receive partially global information. The cloud server has higher credibility but it cannot get all local information from a global view. Therefore, crowdsourcing-based distributed optimization is weak centralized.

iii) **Time-varying topology**. Workers are randomly moving and neighbors in the communication region are randomly changing. Therefore, the connected edges are randomly time-varying.

iv) **Data sensitivity**. Workers are sensitive on data protection. Their own decision variable vectors are kept secret unless the cloud server asks them share for evolution.

## 3 CEC

### 3.1 The Framework of CEC

The framework of CEC is shown in Fig. 3 and pseudocodes are shown in Algorithm 1. CEC contains two kinds of roles, the cloud server $S$ and worker agents $W$. Once receiving $P$, $S$ dispatches it to $W$. Then, $W$ make an initialization in the search space. They communicate uncertainty fitness values with their connected neighbors. After that, comparison results of uncertainty fitness are sent to the cloud server that conducts the competition ranking over comparison results. Next, $S$ makes the uncertainty detection to detect unreliable agents which have high uncertainty levels. If the termination conditions are not reached, $S$ uses ranking and detection results to guide the evolution and update of $W$. Otherwise, the candidate in the top rank is submitted to $R$ as the optimized solution. The detailed process is introduced as follows.

**Step 1)** Task Dispatch. $S$ dispatches the optimization task $P$ with the objective $f(X)$ and search space $D = \{D_1, D_2, ..., D_m\} = \{[lb_1, ub_1], ..., [lb_m, ub_m]\}$ to $n$ worker agents.

**Step 2)** Initialization. Each worker agent randomly generates one individual $X_i$ in $D$. To simulate the environmental-aware uncertainty, each agent attaches an interval uniform random number $r_i = U(b_i, e_i)$ on the fitness evaluation $f(X_i)$. $i$ is the index of the $i^{th}$ agent. $U$ represents to generate a uniform distributed value in interval with the

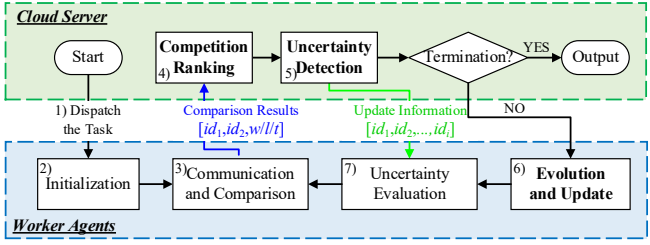

Fig. 3. The framework of CEC.

**Algorithm 1** *Pseudo Codes of CEC*
**Input**: The optimization task $P$ with the objective $f(X)$ and the search space $D$
　　/* Initialization */
1. Randomly initialize one individual $X_i \in D$ on each worker
2. Evaluate $F(X_i) = f(X_i) + r_i$
3. Randomly initialize the topology
4. Evolution generation $g = 1$, exhausted evaluations $fes = 0$
5. **While** fitness evaluations are not exhausted **do**
　　/* Communication */
6. 　**For** each worker $i$
7. 　　**If** $j$ is the neighbor of $i$
8. 　　　Send $F(X_i)$ to $j$
9. 　　　Compare $F(X_i)$ and $F(X_j)$
10. 　　　Send comparison result $[id_1, id_2, w/l/t]$ to $S$
11. 　　**End If**
12. 　**End For**
　　/* Competition Ranking */
13. 　Competition rank on $S$ as Eq. (1)-(5)
　　/* Uncertainty Detection */
14. 　If $mod(g, u) = 0$
15. 　　**If** $rank_i^j = 1$ or $rank_i^j = 4$ for all $j = g - u + 1 : g$
16. 　　　$i$ is detected unreliable and removed from the crowd
17. 　　**End if**
18. 　**End if**
　　/* Communication and Evolution */
19. 　Ranking information is sent by $S$ to workers
20. 　Worker update themselves from higher ranking neighbors through specific evolutionary operators
　　/* Uncertainty Evaluation */
21. 　Evaluate $F(X_i) = f(X_i) + r_i$
22. 　Update $fes$ and $g = g + 1$
23. 　Randomly vary the topology
24. **End while**
**Output**: The optimized solution $X$

beginning $b_i$ and ending $e_i$. Then, all agents evaluate their individuals by the environmental-aware fitness evaluation $F(X_i) = f(X_i) + r_i$. The topology is randomly initialized. (Lines 1 to 4)

**Step 3) Communication.** Worker agents need to communicate with not only neighbors, but also $S$. Firstly, agents can communicate with their connected neighbors. Due to the data sensitivity, agents are unwilling to disclose their decision variable vectors unless they are requested by $S$. Each agent $a_i$ sends the uncertainty fitness $F(X_i)$ to each $j \in N(i)$, where $N(i)$ is the neighbor set of $a_i$. Secondly, agents can communicate with $S$ by sending the tuple of comparison results $[id_1, id_2, w/l/t]$. Specifically, $id_1$ and $id_2$ are the indexes of two connected agents, respectively. $w, l, t$ mean agent $id_1$ has better, worse, or same uncertainty fitness with agent $id_2$, separately. Take a minimization optimization where the optima is bigger than 0 for example, for two connected agents $a_i, a_j$, $[i, j, w]$ means $F(X_i) < F(X_j)$, $[i, j, l]$ means $F(X_i) > F(X_j)$, $[i, j, t]$ means $F(X_i) = F(X_j)$. (Lines 6 to 12)

**Step 4) Competition Ranking.** After receiving comparison results, $S$ can form a comparison matrix $\mathbf{M} = \begin{pmatrix} p_{11} & \cdots & p_{1n} \\ \vdots & \ddots & \vdots \\ p_{n1} & \cdots & p_{nn} \end{pmatrix}$. For each agent $a_i$, $p_{ii} = NAN$, and $p_{ij} = NAN$, if $j \notin N(i)$. $NAN$ means that agents do not compare with themselves or unconnected agents. Therefore, $\mathbf{M}$ may be sparse if the topology is sparse. Based on $\mathbf{M}$, $S$ conducts the competition ranking and gets a ranking sequence of agent indexes $\mathbf{Seq}$, in which higher rank is better. The detail of competition ranking is elaborated in the next part. (Line 13)

**Step 5) Uncertainty Detection.** Based on the ranking sequence, $S$ conducts uncertainty detection to detect agents with high level uncertainties. These agents are removed from the worker crowd to save the budget of fitness evaluations. The detail of uncertainty detection is elaborated in Section 3.3. If the terminated condition is reached, which means the budget of fitness evaluations is exhausted, $S$ evaluates the top-ranked candidate after the uncertainty detection. It should be noticed that $S$ has no environmental-aware uncertainty, which gives the real fitness of the task. Otherwise, the algorithm goes to Step 6) to guide agents for further evolution. (Lines 14 to 18)

**Step 6) Communication and Evolution.** After uncertainty detection, $S$ has a general cognition for the reliability of all agents. It can request agents with better fitness to send their decision variables to neighbors with worse fitness. Worse neighbors can update themselves through specific evolution strategies. (Lines 19 to 20)

**Step 7) Uncertainty Evaluation.** After evolution and update, agents who have learned from their better neighbors conduct the uncertainty evaluation and the algorithm continues to Step 3). At the same time, numbers of exhausted evaluations and evolution generations are updated. Because the crowd are mobile, the topology is dynamic. (Lines 21 to 23)

### 3.2 Competition Ranking

Competition ranking is adopted to make a difference among agents only based on $\mathbf{M}$. Actually, pair-neighbor comparison results are extremely similar with competition scenarios in sports or games. In sports or games, it is impossible that all players have chances to compete and the ranking is often calculated through accumulated scores [34]. Inspired by this, the competition ranking is developed based on $\mathbf{M}$. To make a simple mark, we use $w, l, t$ to represent win, lose and tie respectively. Particularly, $w, l, t$ in $\mathbf{M}$ is marked as 1, 0, 0.5 respectively. The procedure of competition ranking is as follows.

1) A fuzzy matrix $\mathbf{M}_f$, an accumulated win times matrix $\mathbf{M}_w$, and an accumulated lose times matrix $\mathbf{M}_l$ are initialized. Specifically, the fuzzy matrix is initialized as $\mathbf{M}_f = \begin{pmatrix} 0.5 & \cdots & 0.5 \\ \vdots & \ddots & \vdots \\ 0.5 & \cdots & 0.5 \end{pmatrix}_{n \times n}$, which means before comparison, there is no information of results and all agents are regarded as the same. $\mathbf{M}_w$ and $\mathbf{M}_l$ are initialized as $\begin{pmatrix} 0 & \cdots & 0 \\ \vdots & \ddots & \vdots \\ 0 & \cdots & 0 \end{pmatrix}_{n \times n}$, which caused by that before comparison, there is no accumulated time for neither win nor lose.

2) Updating $\mathbf{M}_w$ and $\mathbf{M}_l$ according to $\mathbf{M}$. The update rule is as follows.

$$\begin{aligned}
\mathbf{M}_w(i,j) &= \mathbf{M}_w(i,j)+1, && \text{if } \mathbf{M}(i,j)=1 \\
\mathbf{M}_l(i,j) &= \mathbf{M}_l(i,j)+1, && \text{if } \mathbf{M}(i,j)=0 \quad (2)\\
\left.\begin{aligned}\mathbf{M}_w(i,j) &= \mathbf{M}_w(i,j)+0.5\\ \mathbf{M}_l(i,j) &= \mathbf{M}_l(i,j)+0.5\end{aligned}\right\}, && \text{if } \mathbf{M}(i,j)=0.5
\end{aligned}$$

For each $i,j \in [1,2,...,n]$, $\mathbf{M}(i,j)=1$ means $F(X_i) < F(X_j)$, $\mathbf{M}(i,j)=0$ means $F(X_i) > F(X_j)$, $\mathbf{M}(i,j)=0.5$ means $F(X_i) = F(X_j)$, and $\mathbf{M}(i,j)=NAN$ means $i=j$ or $j \notin N(i)$. It can be concluded from (2) that $\mathbf{M}_w + \mathbf{M}_l = \mathbf{M}_t$, where each element in $\mathbf{M}_t$ is the total comparison times. Particularly, $\mathbf{M}_t$ is a symmetric matrix, in which $\mathbf{M}_t(i,j) = \mathbf{M}_t(j,i)$.

3) Update $\mathbf{M}_f$ based on $\mathbf{M}_w$ and $\mathbf{M}_l$. This process adopts the cosine similarity measure proposed in [34], which has been proved that the updated fuzzy matrix has perfect consistency. The update rule is as follows:

$$\mathbf{M}_f(i,j) = \begin{cases} \frac{\mathbf{M}_w(i,j)}{\mathbf{M}_t(i,j)} - \lambda \times 2^{-\mathbf{M}_t(i,j)/\max \mathbf{M}_t}, & \text{if } \mathbf{M}_w(i,j) > \frac{\mathbf{M}_t(i,j)}{2} \\ 0.5, & \text{if } \mathbf{M}_w(i,j) = \frac{\mathbf{M}_t(i,j)}{2} \quad (3)\\ \frac{\mathbf{M}_w(i,j)}{\mathbf{M}_t(i,j)} + \lambda \times 2^{-\mathbf{M}_t(i,j)/\max \mathbf{M}_t}, & \text{if } \mathbf{M}_w(i,j) < \frac{\mathbf{M}_t(i,j)}{2} \end{cases}$$

where $\max \mathbf{M}_t$ is the biggest number in $\mathbf{M}_t$. The fuzzy value $\mathbf{M}_f(i,j)$ is determined by the win times divided by the total times. That means, two agents are compared using modified win rates to avoid the circumstance that agents with the same win rate but different total times have the same fuzzy value. Therefore, $\lambda \times 2^{-\mathbf{M}_t(i,j)/\max \mathbf{M}_t}$ is used.

Then, $\mathbf{M}_f$ is transformed as follows.

$$\mathbf{M}_f(i,j) = \frac{\mathbf{M}_f(i,j)}{1-\mathbf{M}_f(i,j)} \quad (4)$$

To use the cosine similarity measure, $\mathbf{M}_f$ is transformed as follows.

$$\mathbf{M}_f(i,j) = \frac{\mathbf{M}_f(i,j)}{\sqrt{\mathbf{M}_f(*,j)}} \quad (5)$$

where $\mathbf{M}_f(*,j) = \sum_{i=1}^{n} \mathbf{M}_f(i,j)$, is the sum of columns. The priority vector ***PRI*** is calculated as follows.

$$PRI = \frac{\mathbf{M}_f(i,*)}{\mathbf{M}_f(*,*)} \quad (6)$$

where $\mathbf{M}_f(i,*) = \sum_{j=1}^{n} \mathbf{M}_f(i,j)$ and $\mathbf{M}_f(*,*) = \sum_{i=1}^{n} \mathbf{M}_f(i,*)$. The larger value in ***PRI***, the higher rank the worker has.

This is caused by the relationship between $\mathbf{M}_f$ and ***PRI*** when $\mathbf{M}_f$ is perfectly consistent. For any two vectors $\langle \vec{u}, \vec{v} \rangle$, the cosine similarity measure is calculated as $\cos\langle \vec{u}, \vec{v} \rangle = \frac{\vec{u}\cdot\vec{v}}{||\vec{u}||||\vec{v}||}$. It has been proved that $\mathbf{M}_f$ is perfectly consistent when it satisfies the condition $\mathbf{M}_f(i,j) = \frac{PRI(i)}{PRI(i)+PRI(j)}$ [35]. Therefore, the cosine similarity measure between the column vector $\mathbf{M}_f(*,j)$ and ***PRI*** equals to 1 if $\mathbf{M}_f$ is perfectly consistent.

To facilitate understanding, we give a fully connected topology with five workers as a simple example. Let us suppose $\mathbf{M} = \begin{pmatrix} NAN & 0 & 0 & 0 & 1 \\ 1 & NAN & 1 & 1 & 1 \\ 1 & 0 & NAN & 0 & 1 \\ 1 & 0 & 1 & NAN & 1 \\ 0 & 0 & 0 & 0 & NAN \end{pmatrix}$. By calculating $\mathbf{M}_w, \mathbf{M}_l, \mathbf{M}_t$ by (2), $\mathbf{M}_w = \begin{pmatrix} 0 & 0 & 0 & 0 & 1 \\ 1 & 0 & 1 & 1 & 1 \\ 1 & 0 & 0 & 0 & 1 \\ 1 & 0 & 1 & 0 & 1 \\ 0 & 0 & 0 & 0 & 0 \end{pmatrix}$, $\mathbf{M}_l = \begin{pmatrix} 0 & 1 & 1 & 1 & 0 \\ 0 & 0 & 0 & 0 & 0 \\ 0 & 1 & 0 & 1 & 0 \\ 0 & 1 & 0 & 0 & 0 \\ 1 & 1 & 1 & 1 & 0 \end{pmatrix}$, $\mathbf{M}_t = \begin{pmatrix} 0 & 1 & 1 & 1 & 1 \\ 1 & 0 & 1 & 1 & 1 \\ 1 & 1 & 0 & 1 & 1 \\ 1 & 1 & 1 & 0 & 1 \\ 1 & 1 & 1 & 1 & 0 \end{pmatrix}$. Originally, $\mathbf{M}_f = \begin{pmatrix} 0.5 & 0.5 & 0.5 & 0.5 & 0.5 \\ 0.5 & 0.5 & 0.5 & 0.5 & 0.5 \\ 0.5 & 0.5 & 0.5 & 0.5 & 0.5 \\ 0.5 & 0.5 & 0.5 & 0.5 & 0.5 \\ 0.5 & 0.5 & 0.5 & 0.5 & 0.5 \end{pmatrix}$. After calculation of (3), $\mathbf{M}_f = \begin{pmatrix} 0.5 & 0.005 & 0.005 & 0.005 & 0.995 \\ 0.995 & 0.5 & 0.995 & 0.995 & 0.995 \\ 0.995 & 0.005 & 0.5 & 0.005 & 0.995 \\ 0.995 & 0.005 & 0.995 & 0.5 & 0.995 \\ 0.005 & 0.005 & 0.005 & 0.005 & 0.5 \end{pmatrix}$. By calculating (4) and (5), $\mathbf{M}_f = \begin{pmatrix} 0.0029 & 0.005 & 0 & 0 & 0.5 \\ 0.5773 & 0.9999 & 0.7071 & 1 & 0.5 \\ 0.5773 & 0.005 & 0.0036 & 0 & 0.5 \\ 0.5773 & 0.005 & 0.7071 & 0.005 & 0.5 \\ 0 & 0.005 & 0 & 0 & 0.0025 \end{pmatrix}$. Finally, the priority vector ***PRI*** calculated by (6) is $(0.0707, 0.5270, 0.1512, 0.2499, 0.0011)^T$. Therefore, the final ranking result in terms of sorted index is [4,1,3,2,5]. It indicates the sorted index from the best to the worst after ranking, which means the 2nd agent is best and the 5th agent is worst.

### 3.3 Uncertainty Detection

Uncertainty detection is a strategy to identify unreliable agents, which may be caused by uncertainty of agent behavior and heterogeneity of devices. The worser case is that there exist malicious attackers in the crowd. Therefore, uncertainty detection is significant to identify unreliable agents and save fitness evaluations for further evolution.

Uncertainty detection is based on the priority vector ***PRI***. The process is conducted as line 14 to 18 in Algorithm 1 and demonstrated in Fig. 4. Firstly, all worker agents are re-sorted from high priority to low priority according to ***PRI***. Then they are classified into four levels $L1, L2, L3, L4$ with average $\lfloor n/4 \rfloor$ agents in each level, in which $\lfloor n/4 \rfloor$ is the floor function of total $n$ agents divided by 4 levels. Therefore $L1$ has the highest $\lfloor n/4 \rfloor$ agents, $L2$ has the second higher $\lfloor n/4 \rfloor$ agents, $L3$ has the third higher $\lfloor n/4 \rfloor$ agents and $L4$ has the lowest $n - \lfloor n/4 \rfloor \times 3$ agents. It is a normal case that agents are classified into different levels in different generations. This is caused by that the evolution of agents are not controllable since the topology is changing in each generation. On the contrary, if an agent is frequently classified in $L1$ or $L4$ among several generations, it may be an unreliable agent.

The above phenomenon is reasonable. The uncertainty

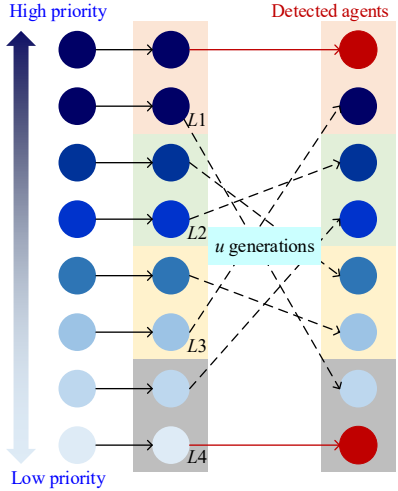

Fig. 4. The illustration of uncertainty detection, in which *L* represents the classified level and *u* is the number of detected generations.

**Algorithm 2** *Pseudo Codes of LLSO*

**Input**: The whole population *P*
1. Divide the population *P* into *L* levels according to the fitness
2. Keep individuals in the first level to the next generation
3. **For** each individual for the 2$^{nd}$ to the 4$^{th}$ level
4.    Randomly select two different higher-level indexes $k1, k2$
5.    **If** it is the 2$^{nd}$ level
6.       $k1 = k2 = 1$
7.    **End If**
8.    Randomly select one individual for $k1, k2, x_{k1,r1}, x_{k2,r2}$
9.    Generate the offspring using the update rule
10.   Check and modify the boundaries of decision boundaries
11. **End For**

**Output**: The updated population

fitness of the *i*th agent $F(X_i)$ is affected by pure fitness $f(X_i)$ and uncertainty $r_i = U(b_i, e_i)$. Although agents only know $F(X_i)$, the relationship between $f(X_i)$ and $r_i$ has great influence on it. Let us take the minimization problem for example. If $|r_i| \ll f(X_i)$, which means the absolute value of $r_i$ is much smaller than $f(X_i)$, $F(X_i)$ is mainly influenced by $f(X_i)$. In this case, the update of $f(X_i)$ may lead to different ranking results as blue circles in Fig. 4. If $|r_i| \gg f(X_i)$, which means the absolute value of $r_i$ is much bigger than $f(X_i)$, $F(X_i)$ is mainly influenced by $r_i$. The update of $f(X_i)$ has little impact on $F(X_i)$. There are two situations of $r_i$. When $r_i$ is a positive number and much bigger than $f(X_i)$, the agent is much worse than other agents and frequently classified in *L*4 as the bottom red circle in Fig. 4. When $r_i$ is a negative number and much smaller than $f(X_i)$, the agent is much better than other agents and frequently classified in *L*1 as the top red circle in Fig. 4.

Therefore, agents which are frequently classified in the first or last class during *u* generations are detected as unreliable ones. They are excluded from the crowd and not optimize the task any more. Here, the number of detected generations *u* is an important parameter, which is set as 100 through experiment investigation.

## 4 EXPERIMENT INVESTIGATION OF CEC

In this section, we firstly introduce the test suites. Secondly, the experiment setting is given, which includes an example of CEC, CLLSO and the parameter setting. Then, we conduct parameter investigation and strategy investigation. Next, experiments on CLLSO and other traditional centralized EC algorithms for large-scale optimization are conducted to illustrate the performance of CEC. Finally, comparisons on distributed clustering optimization between CLLSO and other distributed optimization methods are conducted to illustrate the promising applications of CEC.

### 4.1 Test Suite

To demonstrate the performance of CEC, experiments are conducted on two most widely used large-scale global optimization test suites CEC2010 [36] and CEC2013 [37]. The detailed introduction and implementation codes can be found in their proposed papers. Particularly, CEC2010 has 20 test problems with varies properties and CEC2013 has 15 test problems with more complicated properties. The dimension of problems in CEC2010 and CEC2013 is 1,000, which brings the challenge of the curse of dimensionality.

### 4.2 Experiment Setting

#### 4.2.1 CLLSO

To conduct extensive experiments to illustrate the satisfactory performance of CEC, this part gives an example of CEC, named as CLLSO. CLLSO uses LLSO as the evolutionary optimizer. LLSO is proposed to enhance exploitation and exploration ability of EC for large-scale optimization [33]. The procedures of CLLSO are the same as Fig. 3, where the procedure **Evolution and Update** adopts LLSO evolution process. The pseudocodes of LLSO are shown in Algorithm 2 and procedures are introduced as follows.

In each generation, the swarm *P* is firstly layered into four levels according to fitness. To keep exploitation ability, particles in the first level are directly into the next generation. To improve exploration ability, particles in the other levels can learn from particles in higher levels. In detail, firstly two different higher levels $k1, k2$ are randomly selected. If the current particle $x_i$ is in the 2$^{nd}$ level, $k1 = k2 = 1$. Then one particle is randomly selected from each level to form two learning particles $x_{k1,r1}, x_{k2,r2}$. $x_i$ is updated based on following update rule.

$$v_i^d = r_1 \times x_i^d + r_2 \times (x_{k1,r1}^d - x_i^d) + \varphi \times r_3 \times (x_{k2,r2}^d - x_i^d) \quad (7)$$

$$x_i^d = x_i^d + v_i^d \quad (8)$$

where $v_i^d$ is the velocity of the *i*th particle in *d*th dimension. $r_1, r_2, r_3$ are three random decimals between [0,1]. $\varphi$ is the control parameter and set as 4 in [33]. If any dimensions of the updated particles are out of boundaries, they are reinitialized in the domain.

LLSO has get developments for expensive optimization [38], multi-task optimization [39], fault diagnosis [40] due to its satisfactory performance in large-scale optimization. Therefore, it is adopted as the EC optimizer in this paper to implement an example of CEC.

#### 4.2.2 Parameter Setting

There are three parameters need to be clarified, *FES*, *NP* and *u*. The maximal number of fitness evaluations *FES*

and population size, also known as the crowd size $NP$ are set as suggested in [33]. Specifically, $FES = 1000 \times D$, where $D$ is the dimensionality. $NP = 500$ for problems with 1,000 dimensionalities. New parameter $u$, which is the number of detected generations, is set as 100 through experiment investigation.

To simulate the environmental-aware uncertainty, we also give the uncertainty level setting. To make a difference among uncertainty levels, the exponential form $2^{ul_i}$ is used to set the bound value $bv_i$ of the $i$th agent, where $ul_i$ is the uncertainty level. In the uncertainty detection experiments, we discuss 1) positive uncertainties in which $r_i = U(0, bv_i)$ and 2) negative uncertainties in which $r_i = U(bv_i, 0)$. In the other experiments, we adopt the positive uncertainty $r_i = U(0, bv_i)$.

Besides, experiments are executed on machines with processors Intel® Core™ i5-9400 CPU @2.90GHz and RAM 16.0 GB. All results are averaged over 25 independent runs to make a fair comparison.

### 4.3 Investigation of Competition Ranking

This section investigates the influence of competition ranking. Ranking results totally depend on the comparison matrix, in which the topology sparsity is an important factor. According to ranking results, agents are layered into four levels to conduct level-based learning evolution operator. Therefore, layered accuracy is important for evolution. We set topology sparsity of agents as 0.01, 0.1, 0.2, 0.3, 0.4, 0.5, 0.6, 0.7, 0.8, 0.9, 1.0 for experiments to investigate the layered accuracy from sparsity to density, which means the number of neighbors for each agent equals as $0.01 \times NP$, $0.1 \times NP$, $0.2 \times NP$, $0.3 \times NP$, $0.4 \times NP$, $0.5 \times NP$, $0.6 \times NP$, $0.7 \times NP$, $0.8 \times NP$, $0.9 \times NP$, $1.0 \times NP$. Particularly, $0.01 \times NP$ is a much sparser topology whereas $1.0 \times NP$ is a fully connected topology. Though the topology is time-varying in each generation, the maximal number of neighbors is fixed. Without loss of generality, we take f01 and f03 in CEC2010 as examples to show the layered accuracy for different topology sparsity in Fig. 5 and the optimized results are shown in Fig. 6. Black points are mean layered accuracies of the population during evolution and red lines indicate standard deviations.

It can be concluded from results that sparser topology leads to lower layered accuracy whereas denser topology leads to higher layered accuracy. Specifically, the 100% mean layered accuracy and 0 standard deviation of fully connected topology illustrate the effectiveness of competition ranking strategy. Correspondingly, the optimization performance is becoming better with the topology becoming denser. This phenomenon is easy to understand since sparse topology reflects less information than dense topology. Besides, when the topology sparsity is bigger than 0.1, the layered accuracy is large than 90% and the optimized fitness are stable to the fully connected topology. Therefore, our method is robust to the vary of topology sparsity. The competition ranking on the cloud can well cooperate with the level-based learning operator to lead the evolution of worker agents. To simulate the sparse topology, we set topology sparsity as 0.1 as an example for the following experiments.

### 4.4 Investigation of Uncertainty Detection

This section discusses the uncertainty detection strategy in two cases, 1) positive uncertainties in which $r_i =$

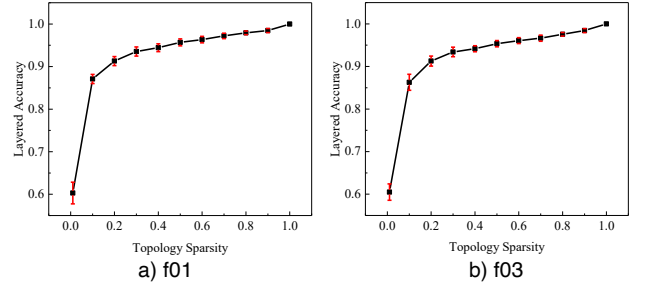

Fig. 5. The layered accuracy for different levels of topology sparsity, in which a) f01, b) f03. Black points are mean layered accuracies of the population during evolution and red lines indicate standard deviations.

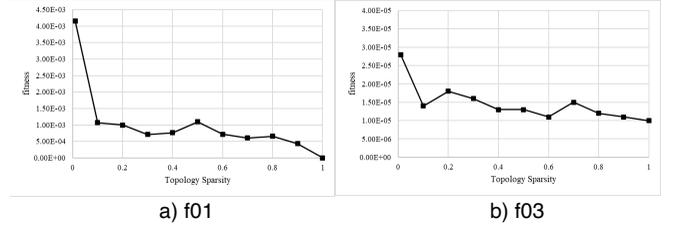

Fig. 6. The optimized fitness for different levels of topology sparsity, in which a) f01, b) f03.

$U(0, bv_i)$ and 2) negative uncertainties in which $r_i = U(bv_i, 0)$. Without loss of generality, we set 10% individuals with uncertainties whose absolute values are bigger than 1. Other 90% individuals with uncertainties whose absolute values are smaller than 1. The maximal value of $|bv_i|$ is set as $2^{30} = 1.0737e9$ out of two reasons. Firstly, the maximal value of $bv_i$ is smaller than the fitness value $f(X_i)$ of randomly initialized individuals. At the beginning of the algorithm, $r_i \ll f(X_i)$ and the uncertainty fitness $F(X_i) = f(X_i) + r_i$ is mainly dominated by $f(X_i)$. Secondly, the maximal value of $bv_i$ is bigger than the global optima fitness value. When the algorithm evolves to the promising areas, $r_i \gg f(X_i)$ and $F(X_i)$ is mainly dominated by $r_i$. Therefore, during the evolution of the algorithm, the relationship between $r_i$ and $f(X_i)$ changes from $r_i \ll f(X_i)$ to $r_i \gg f(X_i)$. It is beneficial to test and observe the algorithm performance in these two circumstances. The specific setting is as follows:

$$|bv_i| = \begin{cases} 2^{-(NP \times 90\% - i)}, & i = 1, 2, \dots, NP \times 90\% \\ 2^{(i - NP \times 90\%) \times \frac{30}{NP \times 10\%}}, & i = NP \times 90\% + 1, \dots, NP \end{cases}$$

As for the time-varying topology, the maximal number of neighbors of each agent is set as $NP \times 0.01$.

#### 4.4.1 Layered Analysis for Uncertainty Detection

This section analyzes layered results for agents with different uncertainty levels during the whole evolution. We take six agents with uncertainty levels $|bv_i| = 6.8791e - 136$, $5.5271e - 76$, $8.8818e - 16$, $64$, $262144$, $1.0737e + 09$ for illustration. For positive uncertainties, $bv_i = |bv_i|$ and $r_i = U(0, bv_i)$. In this circumstance, layered results during evolution are shown in Fig. 7. For negative uncertainties, $bv_i = -|bv_i|$ and $r_i = U(bv_i, 0)$. In this circumstance, layered results during evolution are shown in Fig. 8. We take the average of 10 continuous generations to calculate a smooth layer. Due to the page limit, Fig. 7 and Fig. 8 only show results of f01, f03, f06.

It can be seen from figures that agents with larger uncertainty levels $|bv_i|$ are much easier and earlier to be detected. This is caused by that larger uncertainty value has more influence on ranking results. $F(X_i) = f(X_i) + r_i$ is affected by $f(X_i)$ and $r_i$. If $|r_i|$ is much bigger than $f(X_i)$, $F(X_i)$ is mainly influenced by $r_i$. At the start of evolution, $f(X_i)$ is larger and only extremely high level $|r_i|$ can dominate $F(X_i)$. Therefore, only agents with large uncertainties such as $|bv| = 1.0737E + 09$, are detected. With the population evolution, fitness is becoming better and agents with relatively smaller uncertainties, such as $|bv| = 262144$ or $64$, are detected. Therefore, it is effective for CEC to conduct uncertainty detection and smaller uncertainties can be detected with sufficient evolution.

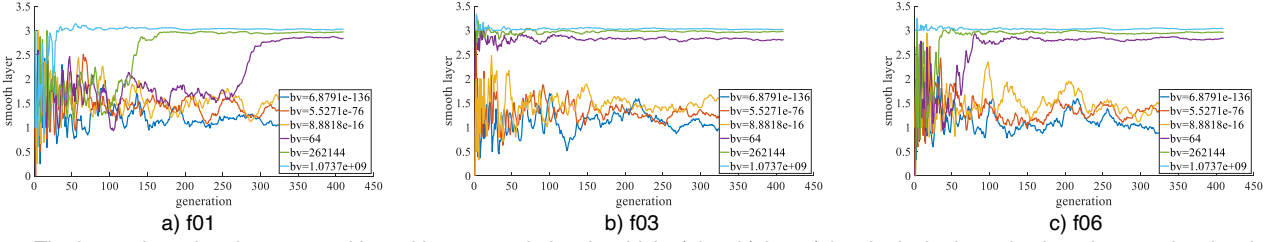

Fig. 7. The layered results of 6 agents with positive uncertainties, in which a) f01, b) f03, c) f06. bv is the bound value of uncertainty level.

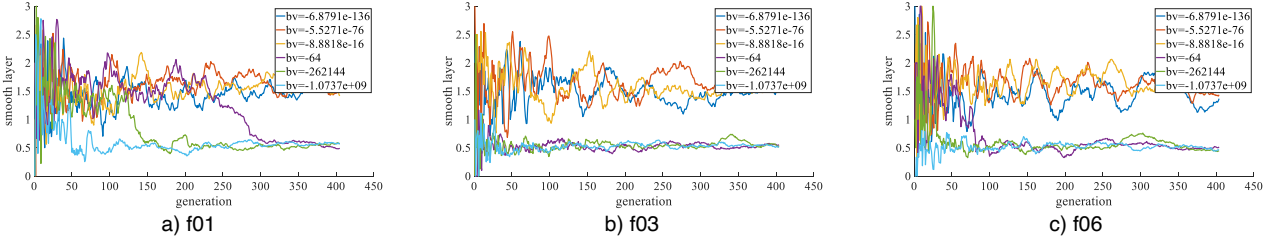

Fig. 8. The layered results of 6 agents with negative uncertainties, in which a) f01, b) f03, c) f06. bv is the bound value of uncertainty level.

### 4.4.2 Investigation of Detection Generations

This section investigates the influence of different numbers of detection generations $u$. We set $u = 50, 100, 150, 200$ and $250$ for experiments. Results of f04, f09, f15 and f18 are shown in Fig. 9.

It can be found from figures that $u = 100$ is a proper setting. Without loss of generality, when $u$ is smaller than 100, the algorithm performance becomes better with the increasing of $u$. The maintenance in the top or bottom level within few generations may be caused by contingencies of better individuals rather than the uncertainty of unreliable ones. Discarding these promising individuals is unfavorable for population evolution. In an extreme case, when $u = 1$, promising individuals are detected and discarded in each generation, which greatly influences the evolution process. When $u$ is bigger than 100, the algorithm performance becomes worse with the increasing of $u$. More detected generations mean longer existence of unreliable individuals. They may mislead the population evolution and exhaust more fitness evaluations. When $u$ equals to the maximal evolution generation, there is no uncertainty detection and the population is totally influenced by individuals with large uncertainties. Therefore, according to experiments, we set the $u = 100$ in our algorithm.

### 4.4.3 Influence Analysis for Uncertainty Detection

To investigate the influence of uncertainty detection, this part conducts experiments to compare the found best solutions between CLLSO with uncertainty detection (CLLSO_wUD) and CLLSO without uncertainty detection (CLLSO_woUD). Results are recorded in Table 2.

It can be concluded that CLLSO_wUD performs better in 16 out of 20 problems than CLLSO_woUD. It is necessary to conduct uncertainty detection out of two reasons. Firstly, individuals with large uncertainties have great influence on population evolution. They may mislead the population to unpromising directions and cannot find the near optima. Secondly, individuals with large uncertainties consume fitness evaluations in each generation. Within a limited budget of fitness evaluations, it would be better to evolve reliable individuals rather than waste evaluations on ones with large uncertainties. Therefore, uncertainty detection is useful to detect individuals with large uncertainties and help the population evolution.

## 4.5 Comparison with Global Optimization Algorithms

Characteristics of crowdsourcing-based distributed optimization, such as uncertainty fitness, weak centralization, time-varying topology and data sensitivity bring great challenges to solve. To illustrate the global optimization ability of CEC, this part conducts experiments to compare the CLLSO with other traditional centralized EC algorithms. Specifically, we take three variants of particle swarm optimizer for large-scale optimization, social-learning particle swarm optimizer (SL-PSO) [41], competitive swarm optimizer (CSO) [42], dynamic level-based learning swarm optimizer (DLLSO) [33] and one cooperative coevolution optimizer for large-scale optimization, differential evolution with cooperative coevolution and differential grouping (DECC-DG) [43]. To make a fair comparison, all parameters are set as recommendations in their proposed papers. The maximal number of fitness evaluations is set as $1000 \times D$. It is worth emphasizing that all compared algorithms are centralized optimization algorithms without any limitations such as environmental-aware uncertainties, data protection or connection topology. In other words, they are tested in traditional global optimization problems rather than distributed optimization problems since it is a challenge to implement these evolution principles with so many limitations. Results of problems with 1,000 dimensionalities in CEC2010 and CEC2013 are shown in Table 3 and Table 4 respectively.

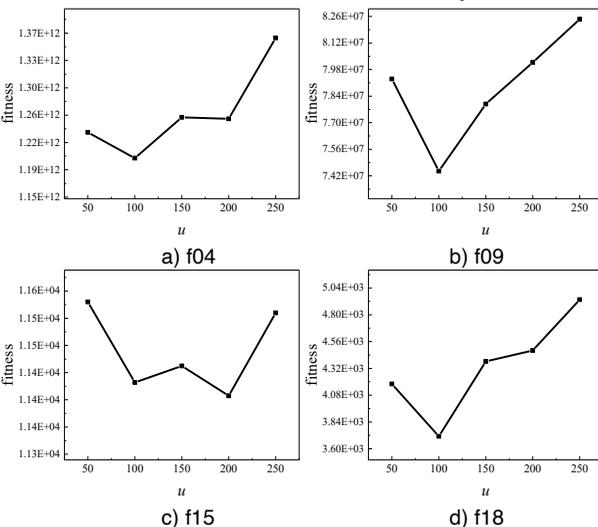

Fig. 9. Performance comparison with different numbers of detection generations $u$, in which a) f04, b) f09, c) f15, d) f18.

TABLE 2
RESULTS COMPARISON BETWEEN CLLSO_WUD AND CLLSO_WOUD

| Problems | CLLSO_woUD mean | std | CLLSO_wUD mean | std |
|---|---|---|---|---|
| f01 | 1.0200E-04 | 1.00E-05 | 0.0000E+00 | 0.00E+00 |
| f02 | 1.1241E+04 | 1.06E+02 | 1.1216E+04 | 1.07E+02 |
| f03 | 2.0000E-05 | 1.00E-06 | 0.0000E+00 | 0.00E+00 |
| f04 | 1.5189E+12 | 2.90E+11 | 1.2034E+12 | 2.02E+11 |
| f05 | 3.3006E+08 | 1.18E+07 | 3.2079E+08 | 1.07E+07 |
| f06 | 3.9900E-04 | 3.00E-05 | 0.0000E+00 | 0.00E+00 |
| f07 | 2.4702E+05 | 1.07E+05 | 3.3280E+02 | 3.40E+02 |
| f08 | 3.8162E+07 | 1.59E+05 | 3.5462E+07 | 2.75E+05 |
| f09 | 8.1738E+07 | 3.90E+06 | 7.4450E+07 | 7.08E+06 |
| f10 | 1.1343E+04 | 9.67E+01 | 1.1343E+04 | 9.37E+01 |
| f11 | 5.5500E-04 | 4.90E-05 | 0.0000E+00 | 0.00E+00 |
| f12 | 1.6861E+06 | 9.26E+04 | 9.8556E+04 | 1.03E+04 |
| f13 | 8.5754E+02 | 3.38E+02 | 9.0334E+02 | 2.87E+02 |
| f14 | 3.4154E+08 | 2.01E+07 | 2.7799E+08 | 1.86E+07 |
| f15 | 1.1388E+04 | 9.35E+01 | 1.1401E+04 | 7.26E+01 |
| f16 | 4.1606E-02 | 2.05E-01 | 2.5000E-05 | 7.00E-06 |
| f17 | 3.7611E+06 | 1.27E+05 | 1.8145E+06 | 2.76E+05 |
| f18 | 3.4836E+03 | 2.12E+03 | 3.7119E+03 | 4.86E+03 |
| f19 | 1.0563E+07 | 5.35E+05 | 9.6938E+06 | 6.83E+05 |
| f20 | 1.0482E+03 | 5.11E+01 | 1.0574E+03 | 8.26E+01 |

Besides mean and standard deviation values, Wilcoxon rank sum test is conducted over 25 independent runs to make the statistical significance test. The last row summarizes testing results with the significance 0.05, in which $\#(+/-/\approx)$ means the number of problems that the compared algorithm is significantly better, worse or has no significant difference than CLLSO. The following conclusions can be got from tables.

DECC-DG performs significantly worse in 15 out of 20 CEC2010 problems and 12 out of 15 CEC2013 problems than CLLSO. DECC-DG is a cooperative coevolution algorithm and it uses differential grouping results for evolution. The performance highly depends on groups of decision variables. It cannot access global information as soon as other centralized global optimization algorithms, such as SL-PSO, CSO, DLLSO. Therefore, it cannot find satisfactory results in most problems. SL-PSO performs significantly worse in 15 out of 20 CEC2010 problems and 12 out of 15 CEC2013 problems than CLLSO. SL-PSO is a widely-used PSO since it adds the social recognition part to help the swarm search for promising areas and get rid of local optima. However, there is a lack of local search strategy for large-scale optimization. Therefore, SL-PSO cannot perform well in large-scale optimization problems. CSO performs significantly worse in 16 out of 20 CEC2010 problems and 12 out of 15 CEC2013 problems than CLLSO. CSO is designed for large-scale optimization. It can significantly improve the population diversity to get rid of local optima due to the competitive learning strategy. However, this may lead to slow convergence since there is no leading of local best particle or global best particle. DLLSO performs significantly worse in 17 out of 20 CEC2010 problems and 12 out of 15 CEC2013 problems than CLLSO. DLLSO is motivated by level-based learning in pedagogy. It layers the swarm into several levels and motivates particles in lower levels learning from particles in higher levels. Though exploration and exploitation abilities are improved, it may also have the problem of slow convergence speed.

Though CEC is designed for distributed optimization with many limitations, it still has satisfactory global optimization ability. The found best results are better or competitive than traditional EC algorithms for centralized large-scale optimization. Competition ranking strategy can well integrate with weak centralization and level-based learning evolution for overcoming the data sensitivity and sparse topology. Uncertainty detection strategy can detect agents with large level uncertainties to alleviate the influence of environmental-aware uncertainties. Therefore, CEC still has satisfactory global optimization performance.

TABLE 3
RESULTS COMPARISON BETWEEN CLLSO AND FOUR OTHER ALGORITHMS IN CEC2010

| | DECC-DG | | | SL-PSO | | | CSO | | | DLLSO | | | CLLSO | |
|---|---|---|---|---|---|---|---|---|---|---|---|---|---|---|
| | mean | std | p | mean | std | p | mean | std | p | mean | std | p | mean | std |
| f01 | 6.02E+05 | 6.9E+05 | 9.7E-11 | 3.62E+4 | 5.3E+4 | 9.7E-11 | 3.62E+2 | 4.3E+1 | 9.7E-11 | 2.23E-1 | 8.9E-2 | 9.7E-11 | 0.00E+0 | 0.0E+0 |
| f02 | 4.41E+03 | 1.4E+02 | 1.4E-09 | 9.25E+3 | 3.9E+2 | 1.4E-09 | 9.38E+3 | 1.2E+2 | 1.4E-09 | 8.40E+3 | 2.8E+2 | 1.4E-09 | 1.12E+4 | 1.1E+2 |
| f03 | 1.66E+01 | 3.9E-01 | 9.7E-11 | 2.11E+1 | 5.5E-2 | 9.7E-11 | 4.53E-2 | 1.6E-2 | 9.7E-11 | 3.38E-4 | 7.0E-5 | 9.7E-11 | 0.00E+0 | 0.0E+0 |
| f04 | 3.24E+13 | 8.9E+12 | 1.4E-09 | 2.34E+6 | 1.1E+6 | 1.4E-09 | 6.1E+12 | 6.E+11 | 1.4E-09 | 2.6E+12 | 5.1E+11 | 1.5E-08 | 1.2E+12 | 2.E+11 |
| f05 | 2.08E+08 | 2.8E+07 | 1.4E-09 | 4.95E+6 | 2.7E+6 | 1.4E-09 | 4.15E+6 | 1.4E+6 | 1.4E-09 | 2.94E+8 | 1.1E+7 | 1.1E-07 | 3.21E+8 | 1.1E+7 |
| f06 | 1.66E+01 | 3.8E-01 | 1.2E-09 | 2.11E+7 | 3.1E+4 | 1.2E-09 | 4.21E+6 | 8.0E+6 | 1.2E-09 | 4.37E-2 | 4.7E-2 | 1.2E-09 | 0.00E+0 | 0.0E+0 |
| f07 | 4.15E+07 | 1.4E+07 | 1.4E-09 | 7.75E+7 | 2.6E+7 | 1.4E-09 | 2.93E+6 | 6.9E+5 | 1.4E-09 | 1.24E+5 | 3.1E+4 | 1.4E-09 | 3.33E+2 | 3.4E+2 |
| f08 | 5.89E+07 | 3.0E+07 | 1.4E-09 | 2.52E+8 | 2.8E+8 | 1.4E-09 | 1.02E+8 | 1.0E+8 | 1.4E-09 | 4.62E+7 | 1.2E+7 | 1.4E-09 | 3.55E+7 | 2.8E+5 |
| f09 | 3.47E+08 | 2.5E+07 | 1.4E-09 | 6.65E+6 | 9.5E+6 | 1.4E-09 | 2.93E+8 | 2.3E+7 | 1.4E-09 | 1.58E+8 | 1.5E+7 | 1.4E-09 | 7.45E+7 | 7.1E+6 |
| f10 | 7.34E+03 | 1.3E+02 | 1.4E-09 | 1.26E+4 | 4.3E+2 | 5.9E-09 | 9.95E+3 | 7.6E+1 | 1.4E-09 | 1.01E+4 | 8.6E+1 | 1.4E-09 | 1.13E+4 | 9.4E+1 |
| f11 | 1.21E+01 | 7.1E-01 | 1.3E-09 | 2.37E+2 | 7.7E-2 | 1.3E-09 | 5.01E+1 | 2.4E+1 | 1.3E-09 | 6.26E-1 | 1.8E+0 | 1.3E-09 | 0.00E+0 | 0.0E+0 |
| f12 | 1.18E+05 | 7.3E+03 | 1.4E-01 | 1.56E+6 | 1.1E+5 | 1.4E-09 | 1.62E+6 | 9.6E+4 | 1.4E-09 | 5.34E+5 | 6.6E+4 | 1.4E-09 | 9.86E+4 | 1.0E+4 |
| f13 | 1.98E+08 | 5.4E+08 | 1.4E-09 | 1.36E+4 | 9.8E+3 | 1.4E-09 | 3.34E+3 | 3.5E+3 | 2.2E-06 | 1.55E+8 | 5.7E+2 | 5.6E-06 | 9.03E+2 | 2.9E+2 |
| f14 | 1.27E+09 | 6.5E+07 | 1.4E-09 | 2.94E+8 | 3.3E+8 | 4.3E-06 | 1.27E+9 | 6.5E+7 | 1.4E-09 | 5.39E+8 | 4.3E+7 | 1.4E-09 | 2.78E+8 | 1.9E07 |
| f15 | 7.22E+03 | 8.9E+01 | 1.4E-09 | 1.12E+4 | 1.2E+2 | 4.6E-09 | 1.02E+4 | 5.2E+1 | 1.4E-09 | 1.04E+4 | 5.7E+1 | 1.4E-09 | 1.14E+4 | 7.3E+1 |
| f16 | 1.83E-02 | 1.3E-03 | 3.7E-07 | 4.32E+2 | 1.1E-1 | 1.4E-09 | 6.89E+0 | 7.6E+0 | 1.4E-09 | 2.93E+0 | 2.6E+0 | 3.2E-09 | 2.50E-5 | 7.0E-6 |
| f17 | 4.27E+06 | 1.8E+04 | 1.4E-09 | 3.47E+6 | 1.5E+5 | 1.4E-09 | 3.45E+6 | 1.7E+5 | 1.4E-09 | 1.99E+6 | 9.2E+4 | 3.5E-08 | 1.81E+6 | 2.8E+5 |
| f18 | 9.69E+10 | 9.3E+09 | 1.4E-09 | 7.17E+4 | 1.8E+4 | 1.4E-09 | 3.97E+4 | 1.1E+4 | 1.4E-09 | 3.75E+4 | 1.0E+4 | 1.4E-09 | 3.71E+3 | 4.9E+3 |
| f19 | 9.86E+06 | 1.7E+05 | 1.4E-09 | 1.27E+7 | 9.2E+5 | 2.0E-09 | 1.12E+7 | 5.6E+5 | 9.3E-09 | 1.15E+7 | 6.1E+5 | 4.1E-09 | 9.69E+6 | 6.8E+5 |
| f20 | 8.22E+09 | 1.6E+09 | 1.4E-09 | 1.10E+4 | 1.6E+3 | 1.4E-09 | 7.32E+3 | 6.2E+2 | 1.4E-09 | 2.69E+3 | 3.6E+2 | 1.4E-09 | 1.06E+3 | 8.3E+1 |
| $\#(+/-/\approx)$=4/15/1 | | | | $\#(+/-/\approx)$=5/15/0 | | | $\#(+/-/\approx)$=4/16/0 | | | $\#(+/-/\approx)$=3/17/0 | | | | | |

TABLE 4
RESULTS COMPARISON BETWEEN CLLSO AND FOUR OTHER ALGORITHMS IN CEC2013

|  | DECC-DG | | | SL-PSO | | | CSO | | | DLLSO | | | CLLSO | |
| --- | --- | --- | --- | --- | --- | --- | --- | --- | --- | --- | --- | --- | --- | --- |
|  | mean | std | p | mean | std | p | mean | std | p | mean | std | p | mean | std |
| f01 | 1.64E+6 | 3.0E+6 | 2.5E-10 | 4.81E+4 | 7.8E+4 | 2.5E-10 | 4.02E+2 | 6.8E+1 | 2.5E-10 | 4.99E-1 | 2.5E-1 | 2.5E-10 | 2.00E-6 | 0.0E+0 |
| f02 | 1.26E+4 | 5.2E+2 | 1.4E-09 | 1.02E+4 | 4.0E+2 | 6.5E-02 | 9.82E+3 | 1.1E+2 | 1.8E-07 | 8.23E+3 | 4.3E+2 | 1.4E-09 | 1.04E+4 | 3.9E+2 |
| f03 | 1.69E+1 | 3.1E-1 | 9.7E-11 | 2.10E+1 | 8.5E-2 | 9.7E-11 | 5.33E+0 | 5.1E+0 | 9.7E-11 | 3.49E-4 | 5.9E-5 | 9.7E-11 | 0.00E+0 | 0.0E+0 |
| f04 | 2.5E+11 | 1.E+11 | 1.4E-09 | 3.7E+10 | 7.0E+9 | 1.4E-09 | 2.6E+10 | 4.4E+9 | 1.4E-09 | 2.6E+10 | 6.7E+9 | 1.4E-09 | 1.1E+10 | 1.7E+9 |
| f05 | 6.97E+5 | 4.0E+5 | 5.2E-09 | 6.19E+6 | 1.8E+6 | 1.3E-04 | 8.98E+5 | 1.6E+5 | 1.4E-09 | 8.03E+5 | 1.3E+6 | 1.4E-09 | 7.93E+6 | 2.0E+5 |
| f06 | 1.86E+4 | 3.1E+4 | 1.4E-09 | 1.06E+6 | 1.7E+3 | 1.4E-09 | 1.67E+1 | 4.9E+0 | 1.4E-09 | 4.86E+0 | 3.3E+0 | 1.4E-09 | 6.21E-3 | 2.0E-2 |
| f07 | 1.31E+9 | 4.0E+8 | 1.4E-09 | 2.53E+9 | 8.6E+8 | 1.4E-09 | 4.05E+8 | 2.3E+8 | 1.4E-09 | 3.67E+7 | 1.3E+7 | 5.1E-06 | 1.74E+7 | 1.6E+7 |
| f08 | 9.3E+15 | 5.E+15 | 1.4E-09 | 7.6E+15 | 1.E+15 | 1.4E-09 | 3.5E+15 | 5.E+14 | 1.4E-09 | 4.2E+14 | 9.E+13 | 1.4E-09 | 1.5E+14 | 3.E+13 |
| f09 | 5.97E+8 | 3.5E+7 | 5.5E-02 | 5.95E+8 | 2.0E+8 | 9.8E-01 | 7.95E+7 | 1.8E+7 | 1.4E-09 | 5.59E+7 | 2.5E+7 | 1.4E-09 | 5.78E+8 | 3.3E+7 |
| f10 | 2.37E+1 | 1.4E+0 | 1.4E-09 | 9.00E+7 | 1.9E+7 | 1.4E-09 | 8.46E+2 | 3.8E+2 | 1.4E-09 | 1.60E+2 | 3.8E+1 | 1.4E-09 | 6.69E-3 | 3.4E-3 |
| f11 | 3.1E+11 | 1.E+11 | 1.4E-09 | 3.4E+11 | 6.E+10 | 1.4E-09 | 7.9E+10 | 4.E+10 | 1.4E-09 | 5.66E+9 | 5.2E+9 | 2.3E-09 | 1.10E+9 | 3.9E+8 |
| f12 | 6.4E+11 | 4.E+10 | 1.4E-09 | 1.28E+4 | 2.9E+3 | 1.4E-09 | 7.65E+3 | 7.5E+2 | 1.4E-09 | 3.21E+3 | 1.3E+3 | 1.4E-09 | 1.16E+3 | 1.3E+2 |
| f13 | 5.3E+10 | 8.9E+9 | 1.4E-09 | 3.3E+10 | 7.3E+9 | 1.4E-09 | 1.9E+10 | 3.8E+9 | 1.4E-09 | 3.25E+9 | 9.9E+8 | 2.4E-06 | 1.63E+9 | 8.8E+8 |
| f14 | 9.9E+10 | 3.2E+10 | 1.4E-09 | 5.4E+11 | 1.E+11 | 1.4E-09 | 3.2E+10 | 1.E+10 | 1.4E-09 | 8.41E+9 | 3.2E+9 | 1.4E-09 | 1.53E+8 | 7.6E+7 |
| f15 | 2.78E+7 | 1.6E+7 | 1.9E-07 | 6.43E+7 | 1.2E+7 | 2.4E-01 | 9.51E+7 | 8.1E+7 | 5.2E-09 | 1.25E+8 | 1.6E+7 | 1.4E-09 | 6.66E+7 | 7.1E+6 |
| #(+/−/≈)=2/12/1 | | | | #(+/−/≈)=0/12/3 | | | #(+/−/≈)=3/12/0 | | | #(+/−/≈)=3/12/0 | | | | |

To investigate the convergence of CEC for distributed optimization and traditional EC algorithms for centralized large-scale optimization, we take f01, f03, f11, f13 as examples in Fig. 10. Due to the order of magnitude is too large, we take the logarithm value based on $e$ of the convergence fitness value ($\ln(fitness)$). It can be found that CLLSO has faster convergence speed and better found-best solution. Firstly, though agents in CEC have environmental-aware uncertainties, they are detected by uncertainty detection strategy to alleviate its negative influence. At the same time, the detected unreliable agents are removed from the crowd, which would not consume evaluations any more. Secondly, with the cooperation of competition ranking and level-based learning evolution strategy, agents can well search for global optima with diversity. Besides, results recorded in Table 3, Table 4 and Fig. 10 also demonstrate that CEC do have good robustness. Though the uncertainty level is fixed for each worker, the real uncertainty value is varied around the uncertainty level for each evaluation. However, CEC has not only better optimized fitness value but also smaller standard deviation. It means that the algorithm is robust to the change of uncertainties. Therefore, CEC has satisfactory global optimization and robustness ability.

### 4.6 Comparison on Distributed Clustering Optimization

It is an increasing demand for optimization in wireless sensor networks (WSNs) after data are collected by distributed sensors. As a consequence, novel problems are arising to be solved and distributed clustering optimization is a representative one. Therefore, we take distributed clustering optimization as a real example to illustrate the application of CEC, which has wide applications in scene segmentation, monitoring application, and so on [44].

Without loss of generality, we take a dataset, UrbanGB, from UCL machine learning repository (https://archive.ics.uci.edu/ml/index.php). It has coordinates, which include longitude [-5.55599, 1.75834] and latitude [50.0797, 57.6956], of 360,177 road accidents occurred in urban areas in Great Britain. These coordinates have totally 469 labelled clustering centers. Though the coordinates of road accidents are known in the dataset, we can regard that they are collected by different sensing workers located on different urban areas in Great Britain, and there is a cloud server to coordinate all workers. In this circumstance, the clustering optimization in UrbanGB can be regarded as a crowdsourcing-based distributed optimization problem and used to test the performance of CEC.

Due to the high computational burden, we use the first 10,000 coordinates and 100 most frequently labelled clustering centers as testing examples. These data are shown in Fig. 11, in which black circles are coordinates and red stars are labelled centers.

Specifically, decision variable vectors are coordinates of clustering center. Considering computational burden, we also set the number of clustering centers to be optimized as 100. The objective is to minimize the within-cluster sum of squares (WCSS), which is measured by the following calculation.

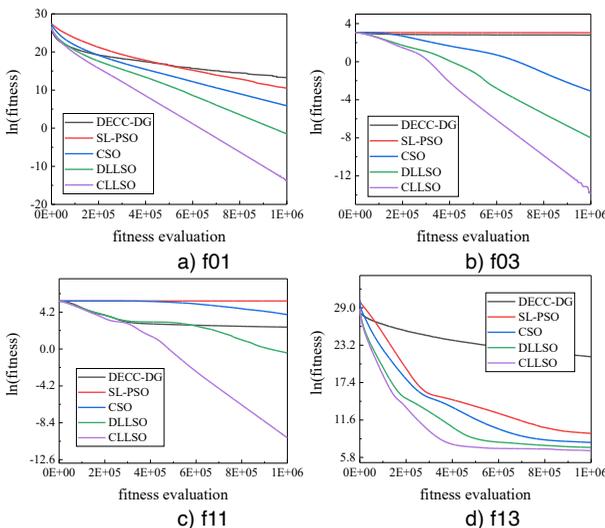

Fig. 10. Comparison of convergence speed, in which a) f01, b) f03, c) f11, e) f13.

$$\min \sum_{j=1}^{k} \sum_{D_i \in j} \| D_i - C_j \|^2 \qquad (9)$$

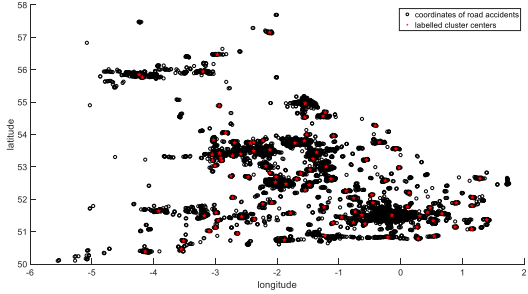

Fig. 11. Illustration of 10,000 coordinates and 100 most frequently labelled clustering centers. Black circles are coordinates of road accidents and red stars are labelled cluster centers.

TABLE 5
RESULTS COMPARISON BETWEEN CLLSO AND TWO DISTRIBUTED CLUSTERING OPTIMIZATION ALGORITHMS

|      | DKM      | MNID     | CLLSO    |
| ---- | -------- | -------- | -------- |
| mean | 1033.154 | 1277.831 | 977.2407 |
| std  | 26.32203 | 157.8187 | 14.39224 |
| p    | 2.64E-09 | 9.73E-11 | —        |

where $k$ is the number of clusters and set as 100 in experiments. It means the optimized dimensionality is 200 and the maximal number of fitness evaluations is 200,000. $C_j$ is coordinates of the $j$th cluster. $D_i$ are coordinate data to be clustered and $D_i \in j$ means $D_i$ has the closest distance to center $j$. WCSS of 100 most frequently labelled clustering centers is 695.7181.

There are many ways to simulate environmental-aware uncertainties in distributed clustering optimization. For example, each worker has mutually different collected data, each worker has partially missing data, each worker has partially additional data, etc. In our experiments, we use the replacement method that the $i$th agent has $i$ uncertain points to randomly replace $i$ original data. To illustrate promising applications of CEC, we take two distributed clustering optimization algorithms for comparison, which are distributed k-means algorithm (DKM) [44] and minimum normalized information distance-based (MNID) [17]. They are implemented and set as their proposed papers. Comparison results are recorded in Table 5. It can be found that CLLSO has lower WCSS value. Besides, Wilcoxon rank-sum test at significance level 0.05 shows that CLLSO performs significantly better than the other compared algorithms. Therefore, CEC has promising applications in both industry and academia.

## 5 CONCLUSIONS

This paper is aware of one kind of emerging distributed optimization led by crowdsourcing and big data era, which is introduced as crowdsourcing-based distributed optimization. Considering its characteristics, this paper combines crowdsourcing paradigm with EC to propose the crowdsourcing-based EC (CEC). With the cooperation of competition ranking and uncertainty detection, CEC can well alleviate challenges such as environmental-aware uncertainties, weak centralization, time-vary topology and data sensitivity. An example, CLLSO is implemented for extensive experiments on benchmark functions and distributed clustering optimization. Results show that CEC has satisfactory performance and promising applications for crowdsourcing-based distributed optimization.

For potential future research, it is worth to study crowdsourcing-based expensive optimization. Explosive data in big data era severely increase computational burden in fitness evaluations. How to improve efficiency within such large volume data is significant for distributed optimization.

## ACKNOWLEDGMENT

This work was supported in part by the National Key Research and Development Project, Ministry of Science and Technology, China, under Grant 2018AAA0101300; in part by the National Natural Science Foundation of China under Grants 61976093 and 61873097; in part by the National Research Foundation of Korea under Grant NRF-2021H1D3A2A01082705; in part by the Guangdong-Shenzhen Joint Key Project No. 2022B1515120076; and in part by the Guangdong Natural Science Foundation Research Team under Grant 2018B030312003. (*Corresponding Author: Wei-Neng Chen*)

## REFERENCES

[1] D. C. Brabham, "Crowdsourcing as a model for problem solving: An introduction and cases," *Converg. Int. Res. into New Media Technol.*, vol. 14, no. 1, pp. 75–90, 2008, doi: 10.1177/1354856507084420.
[2] L. Wang *et al.*, "Task Scheduling in Three-Dimensional Spatial Crowdsourcing: A Social Welfare Perspective," *IEEE Trans. Mob. Comput.*, vol. 14, no. 8, pp. 1–14, 2015, doi: 10.1109/tmc.2022.3175305.
[3] S. S. Bhatti, J. Fan, K. Wang, X. Gao, F. Wu, and G. Chen, "An Approximation Algorithm for Bounded Task Assignment Problem in Spatial Crowdsourcing," *IEEE Trans. Mob. Comput.*, vol. 20, no. 8, pp. 2536–2549, 2021, doi: 10.1109/TMC.2020.2984380.
[4] Y. Xu, M. Xiao, J. Wu, S. Zhang, and G. Gao, "Incentive Mechanism for Spatial Crowdsourcing with Unknown Social-Aware Workers: A Three-Stage Stackelberg Game Approach," *IEEE Trans. Mob. Comput.*, pp. 1–14, 2022, doi: 10.1109/TMC.2022.3157687.
[5] W. Wang *et al.*, "Strategic Social Team Crowdsourcing: Forming a Team of Truthful Workers for Crowdsourcing in Social Networks," *IEEE Trans. Mob. Comput.*, vol. 18, no. 6, pp. 1419–1432, 2019, doi: 10.1109/TMC.2018.2860978.
[6] J. Tan, H. Wu, K.-H. Chow, and S.-H. G. Chan, "Implicit Multimodal Crowdsourcing for Joint RF and Geomagnetic Fingerprinting," *IEEE Trans. Mob. Comput.*, pp. 1–14, 2021, doi: 10.1109/tmc.2021.3088268.
[7] B. Zhao, X. Liu, W. N. Chen, W. Liang, X. Zhang, and R. H. Deng, "PRICE: Privacy and Reliability-Aware Real-Time Incentive System for Crowdsensing," *IEEE Internet Things J.*, vol. 8, no. 24, pp. 17584–17595, 2021, doi: 10.1109/JIOT.2021.3081596.
[8] B. Zhao, X. Liu, W. N. Chen, and R. Deng, "CrowdFL: Privacy-Preserving Mobile Crowdsensing System via Federated Learning," *IEEE Trans. Mob. Comput.*, pp. 1–13, 2022, doi: 10.1109/TMC.2022.3157603.
[9] T. Kohler and M. Nickel, "Crowdsourcing business models that last," *J. Bus. Strategy*, vol. 38, no. 2, pp. 25–32, Jan. 2017, doi: 10.1108/JBS-10-2016-0120.
[10] G. D. Saxton, O. Oh, and R. Kishore, "Rules of Crowdsourcing: Models, Issues, and Systems of Control," *Inf. Syst. Manag.*, vol. 30, no. 1, pp. 2–20, 2013, doi: 10.1080/10580530.2013.739883.
[11] L. Da Xu, W. He, and S. Li, "Internet of things in industries: A survey," *IEEE Trans. Ind. Informatics*, vol. 10, no. 4, pp. 2233–2243, 2014, doi: 10.1109/TII.2014.2300753.
[12] C. Roy, S. Swarup Rautaray, and M. Pandey, "Big Data Optimization Techniques: A Survey," *Int. J. Inf. Eng. Electron. Bus.*, vol. 10, no. 4, pp. 41–48, 2018, doi: 10.5815/ijieeb.2018.04.06.
[13] S. Deng, H. Zhao, W. Fang, J. Yin, S. Dustdar, and A. Y. Zomaya,

"Edge Intelligence: The Confluence of Edge Computing and Artificial Intelligence," *IEEE Internet Things J.*, vol. 7, no. 8, pp. 7457–7469, 2020, doi: 10.1109/JIOT.2020.2984887.

[14] B. Guo *et al.*, "Mobile Crowd Sensing and Computing: The Review of an Emerging Human-Powered Sensing Paradigm," *ACM Comput. Surv.*, vol. 48, no. 1, p. 7, 2015.

[15] T. Yang *et al.*, "A survey of distributed optimization," *Annu. Rev. Control*, vol. 47, pp. 278–305, 2019, doi: 10.1016/j.arcontrol.2019.05.006.

[16] Q. Yang, Y. Liu, T. Chen, and Y. Tong, "Federated machine learning: Concept and applications," *ACM Trans. Intell. Syst. Technol.*, vol. 10, no. 2, pp. 1–19, 2019, doi: 10.1145/3298981.

[17] J. Qin, Y. Zhu, and W. Fu, "Distributed Clustering Algorithm in Sensor Networks via Normalized Information Measures," *IEEE Trans. Signal Process.*, vol. 68, pp. 3266–3279, 2020, doi: 10.1109/TSP.2020.2995506.

[18] B. Lorenzo and S. Glisic, "Optimal routing and traffic scheduling for multihop cellular networks using genetic algorithm," *IEEE Trans. Mob. Comput.*, vol. 12, no. 11, pp. 2274–2288, 2013, doi: 10.1109/TMC.2012.204.

[19] D. Whitley, "An overview of evolutionary algorithms: Practical issues and common pitfalls," *Inf. Softw. Technol.*, vol. 43, no. 14, pp. 817–831, 2001, doi: 10.1016/S0950-5849(01)00188-4.

[20] H. Yang, F. Ye, and B. Sikdar, "A swarm-intelligence-based protocol for data acquisition in networks with mobile sinks," *IEEE Trans. Mob. Comput.*, vol. 7, no. 8, pp. 931–945, 2008, doi: 10.1109/TMC.2007.70783.

[21] A. Chakraborty and A. K. Kar, "Swarm intelligence: A review of algorithms," in *Modeling and Optimization in Science and Technologies*, vol. 10, 2017, pp. 475–494.

[22] Y. Jin, H. Wang, T. Chugh, D. Guo, and K. Miettinen, "Data-Driven Evolutionary Optimization: An Overview and Case Studies," *IEEE Trans. Evol. Comput.*, vol. 23, no. 3, pp. 442–458, 2019, doi: 10.1109/TEVC.2018.2869001.

[23] B. Xue, M. Zhang, W. N. Browne, and X. Yao, "A Survey on Evolutionary Computation Approaches to Feature Selection," *IEEE Trans. Evol. Comput.*, vol. 20, no. 4, pp. 606–626, 2016, doi: 10.1109/TEVC.2015.2504420.

[24] K. C. Tan, L. Feng, and M. Jiang, "Evolutionary Transfer Optimization - A New Frontier in Evolutionary Computation Research," *IEEE Comput. Intell. Mag.*, vol. 16, no. 1, pp. 22–33, 2021, doi: 10.1109/MCI.2020.3039066.

[25] J. Xu, Y. Jin, W. Du, and S. Gu, "A federated data-driven evolutionary algorithm," *Knowledge-Based Syst.*, vol. 233, p. 107532, 2021, doi: 10.1016/j.knosys.2021.107532.

[26] J. Xu, Y. Jin, and W. Du, "A federated data-driven evolutionary algorithm for expensive multi-/many-objective optimization," *Complex Intell. Syst.*, vol. 7, no. 6, pp. 3093–3109, 2021, doi: 10.1007/s40747-021-00506-7.

[27] S. Liu, H. Wang, W. Peng, and W. Yao, "A Surrogate-Assisted Evolutionary Feature Selection Algorithm with Parallel Random Grouping for High-Dimensional Classification," *IEEE Trans. Evol. Comput.*, pp. 1–15, 2022, doi: 10.1109/TEVC.2022.3149601.

[28] F.-F. Wei, W.-N. Chen, Q. Li, S.-W. Jeon, and J. Zhang, "Distributed and Expensive Evolutionary Constrained Optimization with On-Demand Evaluation," *IEEE Trans. Evol. Comput.*, pp. 1–1, 2022, doi: 10.1109/tevc.2022.3177936.

[29] X. Guo, W. Chen, F. Wei, W. Mao, X. Hu, and J. Zhang, "Edge-Cloud Co-Evolutionary Algorithms for Distributed Data-Driven Optimization Problems," *IEEE Trans. Cybern.*, pp. 1–14, 2022.

[30] A. Song, W. Chen, Y. Gong, X. Luo, and J. Zhang, "A Divide-and-Conquer Evolutionary Algorithm for Large-Scale Virtual Network Embedding," *IEEE Trans. Evol. Comput.*, vol. 24, no. 3, pp. 566–580, 2020, doi: 10.1109/TEVC.2019.2941824.

[31] Y. J. Gong *et al.*, "Distributed evolutionary algorithms and their models: A survey of the state-of-the-art," *Appl. Soft Comput. J.*, vol. 34, pp. 286–300, 2015, doi: 10.1016/j.asoc.2015.04.061.

[32] B. Zhao, S. Tang, X. Liu, X. Zhang, and W. N. Chen, "IronM: Privacy-preserving reliability estimation of heterogeneous data for mobile crowdsensing," *IEEE Internet Things J.*, vol. 7, no. 6, pp. 5159–5170, 2020, doi: 10.1109/JIOT.2020.2975546.

[33] Q. Yang, W. N. Chen, J. Da Deng, Y. Li, T. Gu, and J. Zhang, "A Level-Based Learning Swarm Optimizer for Large-Scale Optimization," *IEEE Trans. Evol. Comput.*, vol. 22, no. 4, pp. 578–594, 2018, doi: 10.1109/TEVC.2017.2743016.

[34] X. Chao, G. Kou, T. Li, and Y. Peng, "Jie Ke versus AlphaGo: A ranking approach using decision making method for large-scale data with incomplete information," *Eur. J. Oper. Res.*, vol. 265, no. 1, pp. 239–247, 2018, doi: https://doi.org/10.1016/j.ejor.2017.07.030.

[35] Y. M. Wang, Z. P. Fan, and Z. Hua, "A chi-square method for obtaining a priority vector from multiplicative and fuzzy preference relations," *Eur. J. Oper. Res.*, vol. 182, no. 1, pp. 356–366, 2007, doi: 10.1016/j.ejor.2006.07.020.

[36] T. Ke, L. Xiaodong, S. P. N., Y. Zhenyu, and W. Thomas, "Benchmark Functions for the CEC'2010 Special Session and Competition on Large-Scale Global Optimization," 2010. [Online]. Available: http://goanna.cs.rmit.edu.au/~xiaodong/cec13-lsgo/competition/cec2013-lsgo-benchmark-tech-report.pdf.

[37] L. Xiaodong, T. Ke, S. P. N., Y. Zhenyu, and K. Qin, "Benchmark Functions for the CEC'2013 Special Session and Competition on Large-Scale Global Optimization," 2013. [Online]. Available: http://goanna.cs.rmit.edu.au/~xiaodong/cec13-lsgo/competition/cec2013-lsgo-benchmark-tech-report.pdf.

[38] F.-F. Wei *et al.*, "A Classifier-Assisted Level-Based Learning Swarm Optimizer for Expensive Optimization," *IEEE Trans. Evol. Comput.*, vol. 25, no. 2, pp. 219–233, 2021, doi: 10.1109/tevc.2020.3017865.

[39] Z. Tang, M. Gong, M. Xie, H. Li, and A. K. Qin, "Multi-Task Particle Swarm Optimization with Dynamic Neighbor and Level-Based Inter-Task Learning," *IEEE Trans. Emerg. Top. Comput. Intell.*, vol. 6, no. 2, pp. 300–314, 2022, doi: 10.1109/TETCI.2021.3051970.

[40] J. Guo, X. Li, Z. Lao, Y. Luo, J. Wu, and S. Zhang, "Fault diagnosis of industrial robot reducer by an extreme learning machine with a level-based learning swarm optimizer," *Adv. Mech. Eng.*, vol. 13, no. 5, pp. 1–10, 2021, doi: 10.1177/16878140211019540.

[41] R. Cheng and Y. Jin, "A social learning particle swarm optimization algorithm for scalable optimization," *Inf. Sci. (Ny).*, vol. 291, no. C, pp. 43–60, 2015, doi: 10.1016/j.ins.2014.08.039.

[42] R. Cheng and Y. Jin, "A competitive swarm optimizer for large scale optimization," *IEEE Trans. Cybern.*, vol. 45, no. 2, pp. 191–204, 2015, doi: 10.1109/TCYB.2014.2322602.

[43] M. N. Omidvar, Y. Ling, Y. Mei, and X. Yao, "Cooperative co-evolution with differential grouping for large scale global optimization," *IEEE Trans. Evol. Comput.*, vol. 18, no. 3, pp. 378–393, 2014, doi: 10.1109/FSKD.2016.7603157.

[44] J. Qin, W. Fu, H. Gao, and W. X. Zheng, "Distributed k-Means Algorithm and Fuzzy c-Means Algorithm for Sensor Networks Based on Multiagent Consensus Theory," *IEEE Trans. Cybern.*, vol. 47, no. 3, pp. 772–783, 2017, doi: 10.1109/TCYB.2016.2526683.

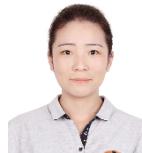

**Feng-Feng Wei** received her Bachelor's degree in computer science from South China University of Technology, Guangzhou, China, in 2019. She is currently a Ph.D student with the School of Computer Science and Engineering, South China University of Technology, Guangzhou and Pazhou Laboratory, Guangzhou, China. Her current research interests include swarm intelligence, evolutionary computation, and their applications on expensive and distributed optimization in real-world problems.

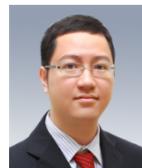

**Wei-Neng Chen** (Senior Member, IEEE) received the bachelor's and Ph.D. degrees in computer science from Sun Yat-sen University, Guangzhou, China, in 2006 and 2012, respectively. Since 2016, he has been a Full Professor with the School of Computer Science and Engineering, South China University of Technology, Guangzhou, and he is now also a professor with Pazhou Lab, Guangzhou, China. He has coauthored over 100 international journal and conference papers, including more than 50 papers published in the IEEE TRANSACTIONS journals. His current research interests include computational intelligence, swarm intelligence, network science, and their applications. Dr. Chen was a recipient of the IEEE Computational Intelligence Society (CIS) Outstanding Dissertation Award in 2016, and the National Science Fund for Excellent Young Scholars in 2016. He is currently the Vice-Chair of the IEEE Guangzhou Section. He is also a Committee Member of the IEEE CIS Emerging Topics Task Force. He serves as an Associate Editor for the IEEE TRANSACTIONS ON NEURAL NETWORKS AND LEARNING SYSTEMS, and the *Complex & Intelligent Systems*.

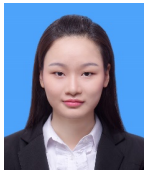

**Xiao-Qi Guo** received the bachelor's degree in computer science and technology from South China University of Technology, Guangzhou, China, in 2018, where she is currently pursuing the Ph.D degree. Her current research interests include evolutionary computation and their applications in distributed networks.

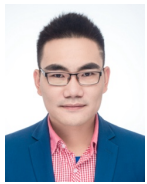

**Bowen Zhao** (Member, IEEE) received his Ph.D. degree in cyberspace security from South China University of Technology, China, in 2020. Now, he is an associate professor at Guangzhou Institute of Technology, Xidian University, Guangzhou, China. His current research interests include privacy-preserving computation and learning and privacy-preserving MCS.

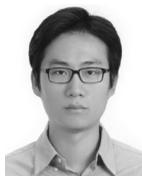

**Sang-Woon Jeon** (Member, IEEE) received the B.S. and M.S. degrees in electrical engineering from Yonsei University, Seoul, South Korea, in 2003 and 2006, respectively, and the Ph.D. degree in electrical engineering from the Korea Advanced Institute of Science and Technology (KAIST), Daejeon, South Korea, in 2011. He has been an Associate Professor with the Department of Military Information Engineering (undergraduate school) and the Department of Electronics and Communication Engineering (graduate school), Hanyang University, Ansan, South Korea, since 2017. From 2011 to 2013, he was a Postdoctoral Associate at the School of Computer and Communication Sciences, Ecole Polytechnique Federale de Lausanne, Lausanne, Switzerland.

From 2013 to 2017, he was an Assistant Professor with the Department of Information and Communication Engineering, Andong National University, Andong, Korea. His research interests include network information theory, wireless communications, sensor networks, and their applications to the Internet of Things and big data. Dr. Jeon was a recipient of the Haedong Young Scholar Award in 2017, which was sponsored by the Haedong Foundation and given by the Korea Institute of Communications and Information Science (KICS), the Best Paper Award of the KICS journals in 2016, the Best Paper Award of the IEEE International Conference on Communications in 2015, the Best Thesis Award from the Department of Electrical Engineering, KAIST, in 2012, the Best Paper Award of the KICS Summer Conference in 2010, and the Bronze Prize of the Samsung Humantech Paper Awards in 2009.

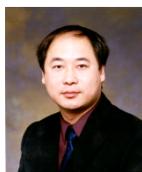

**Jun Zhang** (Fellow, IEEE) received the Ph.D. degree in Electrical Engineering from the City University of Hong Kong in 2002. Currently, he is a visiting professor of Division of Electrical Engineering, Hanyang University. His research interests include computational intelligence, cloud computing, data mining, and power electronic circuits. He has published over 200 technical papers in his research area. Dr. Zhang was a recipient of the China National Funds for Distinguished Young Scientists from the National Natural Science Foundation of China in 2011 and the First-Grade Award in Natural Science Research from the Ministry of Education, China, in 2009. He is currently an Associate Editor of the IEEE TRANSACTIONS ON EVOLUTIONARY COMPUTATION and IEEE TRANSACTIONS ON CYBERNETICS.